# Belief and Surprise - A Belief-Function Formulation

Yen-Teh Hsia
IRIDIA, Université Libre de Bruxelles
50 av. F. Roosevelt, CP 194/6
1050, Brussels, Belgium

## Abstract

We motivate and describe a theory of belief in this paper. This theory is developed with the following view of human belief in mind. Consider the belief that an event E will occur (or has occurred or is occurring). An agent either entertains this belief or does not entertain this belief (i.e., there is no "grade" in entertaining the belief). If the agent chooses to exercise "the will to believe" and entertain this belief, he/she/it is entitled to a degree of confidence c ($1 \geq c > 0$) in doing so. Adopting this view of human belief, we conjecture that whenever an agent entertains the belief that E will occur with c degree of confidence, the agent will be surprised (to the extent c) upon realizing that E did not occur.

## 1 INTRODUCTION

Imagine yourself participating in the following "test". At the beginning of the test, your host announces that an internationally renowned company has built a perfect lottery machine with a very large number of black and white balls inside. This machine is capable of performing the following experiment: emit a ball, and then (perhaps upon user instructions) "retake" the emitted ball. However, the machine has been concealed from you so that you do not see how many balls, black or white, are inside the machine. Twenty one thousand experiments have been run on the machine, and you are given the result.

| Output | Count |
|--------|-------|
| black: | 10424 |
| white: | 10576 |

Given these statistics, you are asked to make a judgment about the ratio of black balls versus white balls. You are not allowed to use pen and pencil (or computers or calculators or ...) to make explicit calculations, and you must make up your mind within a very short time (say, in about thirty seconds). Thus, you are forced to use your own intuition to make the kind of judgment you are asked to make (i.e., just think in ways that you do in your everyday life). Before you know it, your host announces the answer: the actual ratio is 1 (black) versus 1 (white) - i.e., the lottery machine contains N black balls and N white balls, where N may be any natural number (and you will not be informed of its value). Now imagine the extent to which you are surprised by this answer, and let us refer to this "extent of surprise" as 0.

Let us start from scratch and redo the whole test (starting with: at the beginning of the test, your host announces that ...), only this time instead of announcing that the answer is "1 versus 1", your host announces that the actual ratio is "1,000,000,000 versus 1". Again, imagine the extent you are surprised by this answer, and let us refer to this extent of surprise as 10 (actually, 10 should correspond to "x, where x approaches infinity, versus 1"; however, we assume the difference is negligible).

Again, we redo the whole test. Again, the host announces a different answer "x versus y" (e.g., "45 versus 17"), where x > y, and x and y have no common divisor. But this time, you are asked to record your degree of surprise *on a scale of 0 to 10* (with the intuitive meaning of 0 and 10 being what we have just noted above). In doing so, you are not required to use just non-negative integers. Any real number between 0 and 10 can be used.

This, theoretically speaking, allows us to "calibrate" anyone's intuitive degrees of surprise. Once we have made this calibration, we can then use it as a *canonical measurement device* to measure this very same person's degrees of surprise in *any* domain. For example, given the (only) information that the entity we are interested in is a bird, the extent to which we will be surprised by the new information that the entity does not fly may be (judged by us to be) the same as the extent to which we are surprised by the "canonical answer" that the actual ratio is "51 versus 43". If, on a scale of 0 to 10, we recorded the extent of our surprise associated with "51 versus 43" as 4, then the extent to which we will be surprised by the new information that the entity does not fly (given the only information that the entity is a bird) is *measured* 4 (or .4, if we map [0, 10] to [0, 1]). In terms of a notation that we describe in the appendix, we can denote this particular measurement of our intuitive degree of surprise as S([¬FLY] | [BIRD]) = .4. Similarly, if the extent to which we will be surprised by the new information that the entity flies (while previously we were only given the information that it is a bird) is the same as the extent to which we are surprised by the canonical answer that the actual ratio is "1 versus 1", then we can use S([FLY] | [BIRD]) = 0 to denote this measurement. The measurement scheme is clearly subjective, as the extent of surprise associated with "x versus y" (or "not-fly given is-bird") is, in general, different for different people.

Now the reader may be wondering. Why don't we just allow the subject to use pen and pencil (and perhaps even computers) to make whatever calculations he or she feels necessary? Because if the subject is allowed to use probability theory (for example) to make the computations, he or she could have used some probabilistic measure of surprise (e.g., $(\Sigma_j P_j^2)/P_i$, where $P_i$ is the probability of the i-th outcome [Weaver, 1948]) to come up with his or her degree of surprise associated with, say, "51 versus 43". Our answer to this question is as follows. First of all, our purpose here is just to set up an (arguably) useable scheme for the measurement of some one's intuitive degrees of surprise. As far as this purpose is concerned, there is no need to "invoke" the machinery of probability theory here. But even more importantly, what we are trying to do here is to measure some one's intuitive degree of surprise (associated with the occurrence of some event) as what it *is* and not what it *ought to be* (according to some theory). In other words, we want this measurement to be descriptive in some way in characterizing human reasoning. After all, we are entitled to being surprised (according to how we reason and what we actually observe) *without* having to use some formal theory to calculate how surprise we "ought to be".

Why in the world would we want to be able to measure (in a descriptive way) some one's intuitive degrees of surprise? The answer is that we find the following conjecture acceptable.

> **The belief-surprise conjecture:** the *reason* that we are surprised (say, to the extent c > 0) by the occurrence of an event E is that (1) we previously *believed* that E would not occur, and that (2) c was (determined by us to be) the extent to which we were *confident* in entertaining that belief.

Accepting this conjecture, we are led to the following equation:

S("E occurs") = Bel("E does not occur")      (1)

This means, for instance, our beliefs with respect to the above bird-fly example are Bel(([FLY] | [BIRD]) = .4 and Bel(([¬FLY] | [BIRD]) = 0. And here is our thesis: *As we may feel quite comfortable in assessing our intuitive degrees of surprise, we can, in effect, use these measured degrees of surprise to capture our intuitive notion of belief.*

Some philosophical discussions are now in order. Consider the belief that "FLY is true (in the usual propositional sense)." Here, we are adopting the following view of human belief: *An agent either entertains this belief or does not entertain this belief. And when the agent does entertain this belief, he/she/it is entitled to a degree of confidence (c) in doing so.* Thus, for example, by specifying Bel(([FLY] | [BIRD]) = .4, what we mean is that "*given (the truth of) BIRD, we entertain the belief that FLY is true, and .4 is how confident we are in entertaining this belief.*" In other words, we consider the *act* of believing something a categorical action in itself (either we do it, or we do not do it; there is nothing in between), and the uncertainty consists in how confident we are in exercising our "will to believe" and performing that act. This is, of course, only *one* way to think of human belief, as Bel(([FLY] | [BIRD]) = .4 can also be thought of as "given (the truth of) BIRD, .4 is *the extent to which we believe* that FLY is true." In effect, what this latter view of human belief amounts to is a *graded* concept of "entertaining a belief".

We really have no way of telling which of these two views of human belief is more "correct" (there may even be views of human belief that are not stated here). One can only adopt whichever view that looks more natural to him/her. Nevertheless, some may find Equation 1 above questionable, arguing that surprise is actually a *function* of belief. This really depends on which view of human belief one chooses to adopt. Because if the view we advocate here is adopted, then Equation 1 seems perfectly acceptable. But if some other view of human belief is adopted instead, then of course Equation 1 can easily be refuted. So the question is again: *Which view of belief do you find more natural to adopt?* And whichever view we choose to adopt, it may be important to bear in mind the following: *It may be a methodological error if we try to pass judgment on other views of belief from the perspectives of our own view of belief.*

This said, we now need to somehow defend our theory of belief (even before it is presented). As yet another example of our notion of belief, consider the event that "the restaurant run by the Chang family in Ottawa, Kansas, will hire a new waiter or waitress next month." As we know nothing about the recent situation of the restaurant run by the Chang family, we will not be surprised if this event occurs, nor will we be surprised if this event does not occur. In other words, we do not entertain the belief that this event occur (or does not occur), and we can denote it as Bel([HIRE]) = 0 and Bel([¬HIRE]) = 0. Now, this notion of belief is clearly in violation of the additivity axiom of Bayesian probability theory (i.e., Belp([HIRE]) + Belp([¬HIRE]) = 1, where Belp stands for "probabilistic belief"). As the Bayesian theory is well-known for its normative claim, one might quite legitimately wonder whether our yet-to-be-presented theory of belief should even be considered a theory of belief at all.

To address this issue, we need to somehow clarify what the word 'normative' means. According to [Collins, 1987] (a reasonable source of reference for the English language in our view), 'Normative' means *creating or stating norms or rules of behavior*. However, the normative claim of Bayesians may be more than just a claim of "creating or stating norms or rules of behavior." In the words of Ramsey [1931], "(anyone whose subjective belief violates the axioms of probability theory) could have a book (the so-called Dutch Book) made against him by a cunning bettor, and would then stand to lose in any event." In other words, if making an everyday or non-everyday decision is (setting utility considerations aside) like participating in a bet in which *we decide about* the odds for all possible outcomes and *someone else decides about* who bet on what, then the use of a non-probabilistic belief in the decision making process will have the potential of encountering some cunning opponent that make us lose, whatever the actual outcome may be (which, of course, is not what we want). However, the question here is: *Is making an everyday or non-everyday decision like betting in the above described sense (and if so, who may this cunning bettor be)?* This is actually a (very) philosophical question, and no one is obliged to say "yes" (and who this cunning bettor may be) or "no". If we decide to take the stance that decision making is indeed betting in the above described sense, then there is little doubt that our belief ought to be probabilistic (at least we would like it to be the case). But then, what is the reason or rational that we should take this stance? Apart from this philosophical consideration, Bayesian decision making also assumes that our estimation of belief and our estimation of utility can be made *independently*. This makes it even more difficult to accept the normative claim of Bayesians if we find it difficult to do so. It follows that if we do not take the stance that decision making is betting in the above described sense, then as far as Ramsey's argument is concerned, probability theory is no more normative than any other theory of belief, so long as this other theory of belief also serves to create or state norms or rules of behavior.[1]

Our theory of belief is developed with the specific intention of adopting the view of human belief we advocated above. It is a normative theory in the following sense. Given the various "fragments" of belief specified by the user (e.g., Bel([¬STRIKE]) = .3; Bel([RAIN] | [WET]) = .4; Bel([¬RAIN] | [WET]) = 0; Bel([PARTY]) = Bel([PARTY] | [RAIN]),[2] our system will try to infer what the user's belief *must be* under various circumstances.

Here is how the remainder of this paper is organized. In Section 2, we motivate and describe the use of belief functions [Hsia, 1991; Shafer, 1976; Smets, 1988] to characterize our intuitive notion of surprise. The result we get, then, is a theory of belief - a theory that embodies a completely different view of human belief as compared with the view that is embodied in the Bayesian theory of belief. Section 3 gives an illustration of how our theory of belief may be used in a very special context to help people make judgments. Section 4 contains some discussions. Finally, Section 5 concludes.

## 2 BELIEF FUNCTIONS AS A GENERAL FORMALIZATION MECHANISM

Given the canonical measurement device we introduced in the last section, we can now use it to measure our intuitive degrees of surprise associated with any domain (theoretically speaking). But what do we do with the measurements we have made? One answer is that we can "feed" these measurements into a system or a machine and ask it to answer queries like "given these measurements, what can we say (or rather, what can you tell us) about the value of Bel([RAIN] | [PARTY])?" To do that, we need a mathematical formalism that can be used to somehow characterize the measurements we have made - a formalism that serves as a postulate in describing the "inner working mechanism" of our intuitive notion of surprise.

For two reasons, we find the formalism of belief functions [Shafer, 1976; Smets, 1988] - in particular, belief functions in the sense of [Hsia, 1991a][3] - attractive. The first reason is that we think belief functions can be viewed as a very general mechanism that is capable of formalizing various different measurements of surprise. The second reason is that we think the intuitions underlying the belief-function formalism are in line with our intuitive notion of surprise. Let us first talk about the problem of formalizing various measurements of surprise. Consider the following exemplary measurements.

---

[1] Cox [1946] offered another justification for the normative claim of Bayesians. However, Cox has one axiom (about human belief) that we object to - the axiom that our belief in the complement of a proposition is a function of (and thus determined by) our belief in the proposition itself. This axiom is clearly unacceptable, so long as we choose to adopt the view of entertaining beliefs with various degrees of confidence and we also use Equation 1 to measure our degrees of (confidence in entertaining) beliefs. It is easy to find situations in which we are surprised to a different extent by the occurrence of a different event, while being not surprised at all when neither event occurs.

[2] While we use our canonical measurement device to measure the user's degrees of belief, Bayesians may use exchangeable bets to do so.

[3] Hsia [1991a] advocates a "conditioning paradigm" for reasoning with belief functions. In this paradigm, only Dempster's rule of conditioning is used for reasoning, while Dempster's rule of combination is considered something that has to be explicitly justified.

1. Bel([HIRE ∨ ¬HIRE]) = 1

   /* meaning: I will be totally surprised if it turns out that the Changs do not hire any new one, and that they hire some new one. */

2. Bel([HIRE ∧ ¬HIRE]) = 0

   /* meaning: I will not be surprised at all if it turns out that either the Changs do not hire any new one, or that they hire some new one. */

3. Bel([HIRE]) = 0 and Bel([¬HIRE]) = 0

   /* Either way, I will not be surprised. */

4. Bel([Pacifist]) > 0 and Bel([¬Pacifist]) > 0

   /* meaning: I will be surprised if Nixon is not a pacifist (as he is a quaker). I will also be surprised if Nixon is a pacifist (as he is a republican). */

5. Bel([Pacifist]) + Bel([¬Pacifist]) < 1;

   /* meaning: I will only be slightly surprised if Nixon is not a pacifist (as he is also a republican), and I will only be slightly surprised if Nixon is a pacifist (as he is also a quaker). */

6. Bel([TEMP = med ∨ TEMP = low]) > Bel([TEMP = med]) + Bel([TEMP = low]);

   /* meaning: I will not be surprised if the temperature is not medium in the spring (as it can be low), nor will I be surprised if the temperature is not low in the spring (as it can be medium). Nevertheless, I will be surprised if the temperature is high in the spring. */

Viewing these measurements as constraints that are to be satisfied, we can find one or more belief functions (defined below) that satisfy these constraints. However, we cannot use probabilities or even possibilities (in the sense of Zadeh [1978]) to characterize these measurements, as no probability satisfies the third, fifth and sixth constraints and no possibility satisfies the fourth constraint (if Bel is equated with the necessity measure of possibility theory).

One might argue that we should have used Bel([Pacifist] | [R ∧ Q]) instead of Bel([Pacifist]) in the above exemplary measurements. However, the point here is that surprise-measurements such as the ones above *could* happen (in particular, rationality requires that all measurements regarding tautologies and unsatisfiable formulas be in the form of the first and second measurements). With regards to the pacifist example, it does seem more natural to use conditionals. But on the other hand, we would also say that a reasoner should be given total freedom in deciding what ought to be regarded as background information (and stay unspecified in the notation) and what ought to be made explicit (by making it part of the frame (defined below) the reasoner is pondering over).

Our second reason for the choice of belief functions as a formalization mechanism is that the intuitions underlying belief functions are in line with our intuitive notion of surprise. To elaborate, we first need some definitions.

Let $\mathcal{X} = \{X_1, X_2, ..., X_N\}$ be a finite non-empty set of *variables* and let $\Theta_1, \Theta_2, ..., \Theta_N$ be the respective *frames* of these variables (each $\Theta_i$ is a finite non-empty set of values $X_i$ can take; these values are mutually exclusive and exhaustive). $X_i$ is boolean if $\Theta_i = \{Yes, No\}$. $\Theta = \Theta_1 \times \Theta_2 \times ... \times \Theta_N$. We allow the use of *logical formulas* in referring to subsets of $\Theta$, and we list in the appendix the formal correspondence between f, a formula, and [f], f's corresponding subset of $\Theta$.

A *belief function* on $\Theta$ is a function Bel: $2^\Theta \to [0, 1]$ which is characterized by an *m-value function* $m_{Bel}$ (written as "m" whenever confusions can be avoided; m is also called "the m-values of Bel"), where m: $2^\Theta \to [0, 1]$ satisfies two conditions:

(1) $m(\emptyset) = 0$, and
(2) $\sum_{A: A \subseteq \Theta} m(A) = 1$;

and for every subset B of $\Theta$, Bel(B) is defined as $\sum_{A: A \subseteq B} m(A)$.[4] A subset A of $\Theta$ is called a *focal element* of Bel if m(A) > 0. When Bel is such that m($\Theta$) = 1, we call Bel *the vacuous belief function*.

*Dempster's rule of conditioning* is defined as follows. Let Bel be a belief function on $\Theta$ and m be its associated m-values. Let B be a non-empty subset of $\Theta$ such that Bel($B^c$) ≠ 1.

∀ C ⊆ $\Theta$, if C ⊆ B
then m(C | B) df= $\sum_{D: D \subseteq B^c} m(C \cup D) / K$
else m(C | B) df= 0,

where K = 1 - Bel($B^c$) is the normalization constant.

(Note that for every subset S of $\Theta$, Bel(S ∩ B | B) = Bel(S | B), but in general, m(S ∩ B | B) ≠ m(S | B).)

Very abstractly, what the above definition of belief functions says is this. In trying to establish Bel (i.e., to satisfy the specified constraints), we may decide to commit various degrees of intuitive supports (the m-values) to various propositions (i.e., subsets of $\Theta$), and a proposition A is allocated some intuitive support s whenever we find the proposition as a whole deserves this much intuitive support and we do not want to further "split" s among the elements (or rather, subsets) of A. A good example of this is (again) the Chang family example we described in the last section. For simplicity, let us assume we only need to worry about one variable HIRE. Clearly, we want the proposition [HIRE ∨ ¬HIRE] to receive the intuitive support 1. Nevertheless, we do not want to further split this intuitive support among the subsets of [HIRE ∨ ¬HIRE], as our intuitions satisfy Bel([HIRE]) = 0 and Bel([¬HIRE]) = 0 (i.e., either way, we will not be surprised). What the corresponding belief-function formalization suggests, then, is that we intuitively commit the intuitive support 1 to [HIRE ∨ ¬HIRE] and we do not commit any intuitive support to anything else. In practice, this may well be what is happening with our intuitions. Of course, there is no way we can generalize this particular example to all possible situations. Nonetheless, the definition of belief functions serves as a (reasonable) postulate, suggesting that we intuitively *do* commit various intuitive supports to various propositions.

---

[4]This definition is consistent with [Shafer, 1976]. Smets [1988] has a slightly more general definition (called an "open world" definition) in which m($\emptyset$) does not have to be 0 and Bel(A) is defined as the sum of the m-values of those *non-empty* subsets of A.

Given this notion of intuitively committing various intuitive supports to various propositions, Dempster's rule works as follows. Case 1 (C ⊆ B): originally we committed m(C∪D) = s to C∪D, as we considered C∪D as a whole deserved this much (s) intuitive support and we did not want to further "split" s among the subsets of C∪D; now we learn that the actual situation is in B; as a result, we let C "inherit" s, as we *still* consider C as a whole deserves this much intuitive support and we *still* do not want to further "split" s among the subsets of C. Case 2 (C ⊆ $B^c$): originally we considered C the most specific subset of Θ that deserves m(C) = v; now we learn that the actual situation is *not* in $B^c$; as our intuitions satisfy Bel(B | B) = 1 and Bel(C | B) = 0, rationality requires that we make m(C | B) zero and redistribute v in some way; what we do then is that we redistribute v among the focal elements of Bel(. | B) by proportions - a normalization process that is similar in spirit to what the Bayesian rule of conditioning does.

Is this a reasonable concept of conditioning? We think it is. Consider the following example. Suppose we think that one of Tom, Jerry and Pluto broke the window, but are unable to make a further distinction among the three (i.e., we view all three of them as equally likely suspects). However, we are not totally sure about it, as it is also possible that someone else did it. Thus, letting X be the one who broke the window (X=O means "other people did it"), we might specify our belief as Bel([X=T ∨ X=J ∨ X=P ∨ X=O]) = 1, Bel([X=T ∨ X=J ∨ X=P]) = .6, Bel([X=T ∨ X=J]) = 0, Bel([X=T ∨ X=P]) = 0, Bel([X=T]) = 0, etc. The underlying intuition, then, is that (for example) if we later discover that neither Tom nor Jerry broke the window, we will not be surprised (as it then means that Pluto is *the* suspect, and we are happy in being able to isolate the suspect). But if what we subsequently discover is that none of these three broke the window (e.g., we are told by Miss White, their school teacher, that all three of them were cleaning the storage room under her supervision during the time in which the window was broken), then we will surely be surprised (with .6 being the extent of our surprise). With the assumption that the only variable we need to worry about is X, we can characterize the above measurements as the belief function that m([X=T ∨ X=J ∨ X=B ∨ X=O]) = .4 and m([X=T ∨ X=J ∨ X=B]) = .6. In turn, Dempster's rule will give us the expected results. Again, there is no way we can generalize this particular example to all possible situations. Nevertheless, we hope to have convinced the reader in some way that, so far as we are able to tell, Dempster's rule of conditioning seems "compatible" with our intuitive notion of surprise.

Now is a good time to make clear how our notation of belief should be read. By specifying Bel([α] | [β]) = c, where α and β are logical formulas and 1 ≥ c > 0, we mean either (1) or (2) or (3) below, and we do *not* mean either (4) or (5) below.

(1) Given that β is true, I entertain the belief that α is true, and c is how confident I am in entertaining this belief.

(2) Given that β is true, I think α is true, and c is how confident I am in entertaining this belief.

(3) Given that β is true, c is the extent to which I will be surprised upon realizing that α is false.

(4) Given that β is true, c is my belief that α is true (or equivalently, c is my belief in α's being true).

(5) Given that β is true, c is the extent to which I believe that α is true.

We reject (4) as a way to read 'Bel([α] | [β]) = c', because (4) is ambiguous. As such, we feel reading 'Bel([α] | [β]) = c' as (4) has the dangerous potential of inviting some *unintended* view of human belief to "sneak in". This is definitely not how we want our theory of belief to be understood. In the same vein, we reject (5) as a way to read 'Bel([α] | [β]) = c', because we feel (5) really corresponds to the view of human belief that is embodied in the Bayesian theory of belief. For example, if one accepts to read 'Bel([α] | [β]) = c' as (5), then one may want to accept Cox' axioms, while rejecting Equation 1 in the last section. On the other hand, if one accepts to read 'Bel([α] | [β]) = c' as (1) or (2), then one may find Equation 1 perfectly acceptable, while rejecting the one axiom of Cox that we object to in footnote #1.

Similarly, by specifying Bel([α] | [β]) = 0, we mean either (1) or (2) or (3) below.

(1) Given that β is true, I do not entertain the belief that α is true.

(2) Given that β is true, I do not think α is true.

(3) Given that β is true, I will not be surprised (at all) upon realizing that α is false.

We suggest these ways of reading 'Bel([α] | [β]) = 0' because we feel it may be unnatural to assert something like "given that β is true, I entertain the belief that α is true, and I have no confidence whatsoever in doing so."

Having described what belief functions are (from the perspectives of surprise and also from the perspectives of *our* view of human belief), we now need to describe how we can use this formalism for uncertain reasoning. As we have already suggested at the beginning of this section, the basic idea is just to solicit knowledge from the user and then let the system make inferences according to what the belief-function formalism postulates to be the "inner working mechanism of surprise". This amounts to the following two-step reasoning approach.

**Step One** - knowledge solicitation: the user specifies what his or her intuition satisfies. The result is a set of constraints (e.g., the constraints that Bel([¬STRIKE]) = .3, Bel([RAIN] | [WET]) = .4, Bel([¬RAIN] | [WET]) = 0, and Bel([PARTY]) = Bel([PARTY] | [RAIN])).

**Step Two** - reasoning: given the specified constraints, the system then infers properties (e.g., Bel([RAIN] | [PARTY]) ≥ 0) that are satisfied by *all* belief functions satisfying the user-specified constraints.

We acknowledge that this two-step reasoning approach is not as powerful as what we would like it to be. In particular, we have not provided a methodology which, when followed, would allow the user to make a systematic specification of what his or her intuition satisfies. Nevertheless, this reasoning approach serves as the *backbone* of any future, more refined reasoning approach we may wish to devise. There is already some progress in this direction. Hsia [1991a], for example, suggested the use of the principle of minimum commitment[5] on the part of the system to come up with answers like Bel([RAIN] | [PARTY]) = 0 (instead of Bel([RAIN] | [PARTY]) ≥ 0). This allows the system to infer what the user's belief *is* and not what the user's belief *can be* (which is what we are doing here). The principle of minimum commitment is not a panacea, however, as it is not always applicable (i.e., there may not exist a minimum committed belief function in the set of all belief functions satisfying the given constraints). Nevertheless, under this principle, we may be able to devise various specification methodologies that guarantee the existence of a minimum committed belief function satisfying the user-specified constraints. Hsia [1991c] also described a proof theory with belief functions being used in the corresponding model theory. This proof theory is at least as powerful as the system of Pearl and Geffner [1988].

---

[5] A set of belief functions $\mathbb{B}$ has a minimum committed element σ if and only if σ ∈ $\mathbb{B}$ and σ is such that ∀ τ ∈ $\mathbb{B}$, ∀ A ⊆ Θ, σ(A) ≤ τ(A).

## 3 A CASE STUDY

In February of 1991, during the Persian Gulf war, the bombs of the Allied forces hit a bunker in which many civilians were taking shelter. The death toll was high, and one can surely imagine that the Allied forces were greatly surprised by the presence of a large number of civilians inside the bunker. To explain why the bunker was attacked, it was later revealed that there were two pieces of evidence suggesting that the bunker was used for military purposes. One piece of evidence was that satellite photographs showed that military personnels were going into and out of the bunker. The other piece of evidence was that there were military communications between this particular bunker and military installations elsewhere.

Our purpose here is not to pass judgment. We mention this particular incidence only because it happens to be a highly specialized situation in which our theory of belief may be of help to people who may need to decide about their beliefs. Consider the following question. Suppose we are in a situation in which we just obtained the *second* piece of evidence (say, the existence of military communications between the bunker and elsewhere, denoted as "E = Yes"). Should our confidence in entertaining the belief that the bunker is a military bunker be raised significantly? Of course, by asking this question, we are assuming that we have already obtained the first piece of evidence (i.e., satellite photographs showing military personnels going into and out of the bunker, denoted as "P = Yes"), and that based on this first piece of evidence, we have decided that our intuitions satisfy Bel([M] | [P]) = c and Bel([¬M] | [P]) = 0, where M (= Yes) stands for "the bunker is used for military purposes." In other words, given the first piece of evidence, we will not be surprised if it turns out that the bunker is for military purposes, and we will be somewhat surprised (to the extent c) if it turns out that the bunker is a civilian shelter (we assume that a bunker is either a military bunker or a civilian shelter). The same thing can be said about the second piece of evidence. That is, we may decide that our intuitions satisfy Bel([M] | [E]) = d and Bel([¬M] | [E]) = 0. The question, however, is what we should decide about Bel([M] | [P ∧ E]) and Bel([¬M] | [P ∧ E]).

As a first step of the analysis, we let $\mathcal{X}$ = {M, P, E}, and the constraints we have for the moment are the following: Bel([M] | [P]) = c, Bel([¬M] | [P]) = 0, Bel([M] | [E]) = d, and Bel([¬M] | [E]) = 0. Let us now try to add more constraints to this set. First, our intuitions may be such that Bel([M]) = 0, Bel([¬M]) = 0, Bel([P]) = 0, Bel([¬P]) = 0, Bel([E]) = 0, and Bel([¬E]) = 0. That is, we do not entertain *any* belief regarding the bunker (at least we try to be so), whether it is a belief regarding what the bunker is for, a belief regarding whether the satellite photographs will show military personnels going into and out of the bunker, or a belief regarding whether there exists any military communication between the bunker and elsewhere. These "vacuous priors" are, in effect, the kinds of "attitudes" we usually try to enforce upon ourselves when performing evidential reasoning. Adding these six constraints to the original set of constraints, we now have a total of ten constraints that have to be satisfied.

Can we do better (in adding more constraints)? It happens that in this particular case, we can. Because our intuitions are also such that Bel([M ⊃ P]) = 1 and Bel([M ⊃ E]) = 1 are satisfied. So, altogether, we now have twelve constraints to be satisfied: Bel([M] | [P]) = c, Bel(¬M] | [P]) = 0, Bel([M] | [E]) = d, Bel([¬M] | [E]) = 0, Bel([M]) = 0, Bel([¬M]) = 0, Bel([P]) = 0, Bel([¬P]) = 0, Bel([E]) = 0, Bel([¬E]) = 0, Bel([M ⊃ P]) = 1 and Bel([M ⊃ E]) = 1.

Next question is guaranteed to be a thorny one: do our intuitions satisfy Bel([¬P] | [¬M]) = c and Bel([¬E] | [¬M]) = d? In other words, given that the bunker is a civilian shelter, will we be surprised (to the extent c) upon seeing military personnels going into and out of the bunker? Similarly, given that the bunker is a civilian shelter, will we be surprised (to the extent d) upon intercepting military communications between the bunker and elsewhere? To answer these two questions, we need to step back and think: why do our intuitions satisfy Bel([M] | [P]) = c (or Bel([M] | [E]) = d) in the first place? Well, our intuitions satisfy Bel([M] | [P]) = c because, given the only information that there were military personnels going into and out of the bunker, we are confident (to the extent c) in thinking that *other* reasons of why there were military personnels in presence can be *ruled out* (and that the reason of P's being true is because the bunker is for military purposes). Now, *if* upon learning that the bunker is a civilian shelter, we are still this confident (to the extent c) in thinking that other possible causes of the presence of military personnels can be ruled out, *then* we are certainly entitled to the expectation (with c being the corresponding confidence in having the expectation) that there will *not* be military personnels going into and out of the bunker. The same thing can be said about Bel([M] | [E]) and Bel([¬E] | [¬M]). Note that we are not suggesting that contrapositions are *always* satisfied. What we are suggesting is that contrapositions are not all that unreasonable as far as evidential reasoning from the perspectives of surprise is concerned. In fact, for this particular example, we tend to accept contrapositions here. Just think: wouldn't you be surprised (to the extent d) if given the only information that the bunker is a civilian shelter, you later intercept military communications between the bunker and elsewhere? (Remember that you already agreed that, given the only information that you intercepted military communications between the bunker and elsewhere, you will be surprised to the extent d if you later learn that it is a civilian shelter.)

Now back to the example. Adding the two contrapositions (i.e., Bel([¬P] | [¬M]) = c and Bel([¬E] | [¬M]) = d) to our set of constraints, we now have fourteen in the set. If we are able to add two additional independence constraints into this set, then it can be shown [Hsia, 1991b] that, given that both P and E are true, we are entitled to a significantly higher confidence in entertaining the belief that M is true. (In effect, this increase is due to the fact that Dempster's rule of combination *happens to be* what we get when we try to satisfy all sixteen constraints.) But unfortunately, this is where the example fails to satisfy.

The two constraints that are in need here are: Bel([¬P] | [¬M]) = Bel([¬P] | [¬M ∧ E]) and Bel([¬E] | [¬M]) = Bel([¬E] | [¬M ∧ P]). What they mean is as follows. Suppose the bunker is a civilian shelter. Then since we accept contraposition here, we will be surprised (to the extent c) upon seeing military personnels going into and out of the bunker. Nevertheless, suppose the bunker is a civilian shelter and we also intercepted military communications between the bunker and elsewhere, then we (probably) will be less surprised upon seeing military personnels going into and out of the bunker, as something fishy may be going on. The same thing can be said about Bel([¬E] | [¬M]) and Bel([¬E] | [¬M ∧ P]). In other words, our intuitions do not satisfy Bel([¬P] | [¬M]) = Bel([¬P] | [¬M ∧ E]) and Bel([¬E] | [¬M]) = Bel([¬E] | [¬M ∧ P]). Thus, our theory can only stop here, as we lack the necessary ingredients to significantly raise the user's degree of confidence in entertaining the belief that M is true. In some sense, this suggests that we ought not make Bel([M] | [P ∧ E]) significantly higher than Bel([M] | [P]) or Bel([M] | [E]).

## 4 DISCUSSION

When Shafer introduced the theory of belief functions in his 1976 monograph, he had the intention of viewing his theory as a generalization of the Bayesian theory of subjective probability. That is, Shafer's notion of belief is basically what we call *a "graded" concept of entertaining a belief*, and he was, as Fagin and Halpern [1989] observed, extending the Bayesian view of belief from measurable sets to nonmeasurable sets. Smets [1988] also used belief functions to develop a notion of belief along a similar line. Thus, we should not be surprised to see that most of the interpretations of belief functions (e.g., [Black, 1987; Halpern and Fagin, 1990; Kyburg, 1987; Laskey and Lehner, 1989; Nguyen 1978; Pearl, 1988, chapter 9; Shafer and Tversky, 1985]) relate belief functions to probability theory in some way. In other words, all these interpretations share the common goal of trying to generalize the Bayesian view of belief.

Our approach here is completely different. We start with *an entirely different view* of (human) belief, and we "happen" to settle on the use of belief functions to implement (if you will) our particular view of belief. Thus, our theory of belief is (intuitively) *not* a generalization of the Bayesian theory, though formally we cannot deny the fact that probabilities happen to be a special kind of belief functions. In other words, to use our theory of belief for uncertain reasoning, it has to be the case that the user finds our view of belief attractive (and, as a result, wants to adopt it), and it should *not* be the case that the user considers our theory of belief a generalization of the Bayesian theory of belief (as it would then be an outright mistake). So if the question is "why should I use your theory of belief for uncertain reasoning?", then the answer would be "because you agree with us in thinking that the notion of belief consists in entertaining various beliefs with various degrees of confidence."

Up until now, we have maintained that belief functions may be viewed as a very general mechanism for formalizing the notion of surprise. In effect, this just means that we want to keep the versatility of belief functions at our disposal. However, this does not mean that we will always need all the versatility of belief functions when we try to formalize our intuitive notion of surprise associated with some domain. In fact, it may be quite desirable (and also feasible) to impose special constraints when dealing with special domains. For example, in formalizing highly specialized expertise with regards to well-defined domains, we may only need to consider consonant belief functions.[6] And when this is the case, we may then relate belief functions to possibility theory [Dubois and Prade, 1990], as the necessity measure of a possibility is formally equivalent to Bel when Bel is a consonant belief function [Dubois and Prade, 1988]. Similarly, in formalizing common sense, we may only need to consider conjunctive belief functions [Hsia, 1991c].[7] This, in turn, may permit us to make purely qualitative inferences in a logical framework.

In short, belief functions may be "customized" in various ways to permit more efficient specifications as well as inferences. It all depends on whether our domain-of-interest allows us to impose such restrictions.

## 5  CONCLUSION

There is not just *one* view of human belief. There are at least two views: one that embodies a graded concept of believing in something, and the other that embodies the view of entertaining a belief with some degree of confidence. We can of course adopt the view that we always believe in something to a certain extent, in which case we (most likely) would arrive at the Bayesian theory of belief - a theory that has enjoyed a long history of research and development. On the other hand, we can also adopt the view that we entertain beliefs with various degrees of confidence, in which case we would arrive at our current theory of belief. The two views of belief are equally valid (or rather, neither is more "true" than the other). So which view of belief should we adopt? It is, very simply put, anyone's choice.

---

[6] A belief function Bel is consonant if we can arrange the focal elements of Bel in a sequence so that each is contained in the following one.

[7] A belief function Bel is conjunctive if $\forall A \subseteq \Theta, \forall B \subseteq \Theta, \forall C \subseteq \Theta$, if $Bel(A \mid B) > 0$ and $Bel(\Theta \backslash A \mid B) = 0$ and $Bel(C \mid B) > 0$ and $Bel(\Theta \backslash C \mid B) = 0$, then $Bel(A \cap C \mid B) > 0$ and $Bel(\Theta \backslash (A \cap C) \mid B) = 0$.

We set out to develop our theory of belief for the following reason. Surprise is something that has to do with how we reason and what we actually observe in our everyday life. It is an intuitive concept that we *may* feel quite comfortable in assessing its value. Therefore, by adopting the view of belief we advocate in this paper, we can, in effect, capture the notion of belief using our measured degrees of surprise. Ultimately, this may contribute to efforts in the area of "approximating human expertise with the use of computers," also known as artificial intelligence.

In introducing our theory, we have described a canonical measurement device that can be used for the measurement of surprise (and thus belief), and we have suggested the use of belief functions as a very general mechanism for modeling our notion of belief. As an illustration of how our theory of belief may be of some help to people who need to decide about their beliefs, we also gave an example in (boolean and abductive) evidential reasoning. Unlike probability theory which enjoys a wide spectrum of results and applications, our enterprise of surprise and belief is still at the beginning of its development, and we still need to do (much) more work in order to make this enterprise truly accessible to practitioners of uncertain reasoning.

### Acknowledgements

The author thanks Philippe Smets and Paolo Garbolino for ever-enlightening discussions. Robert Kennes and Alessandro Saffiotti also helped to sharpen our views about belief. Thanks also go to one referee who requested that the difference between our theory of belief and the Bayesian theory of belief be clearly specified. This work was supported in part by the DRUMS project funded by the Commission of the European Communities under the ESPRIT II-Program, Basic Research Project 3085.

### Appendix - logical formulas and subsets of $\Theta$

Let $X_1, X_2, ..., X_N$ be variables and $\Theta_1, \Theta_2, ..., \Theta_N$ be their respective frames. $X_i$ is boolean if $\Theta_i = \{Yes, No\}$. Let $\Theta = \Theta_1 \times \Theta_2 \times ... \times \Theta_N$. By the "$X_i$-value" ($1 \leq i \leq N$) of an element $<a_1, a_2, ..., a_N>$ of $\Theta$, we mean $a_i$. Let $x \in \Theta$ and $a \in \Theta_i$ ($1 \leq i \leq N$), we recursively define what a formula f is and whether x *satisfies* the formula f below.

Case 1.   f is "$X_i = a$": x *satisfies* f if and only if the $X_i$-value of x is a. ("$X_i$" is also used as a shorthand for "$X_i = Yes$" in the case of boolean variables.)

Case 2.    f is "¬g", where g is a formula: x *satisfies* f if and only if x does not satisfy g.

Case 3.    f is "g ∨ h", where g and h are formulas: x *satisfies* f if and only if x satisfies at least one of g and h.

Case 4.    f is "g ∧ h", where g and h are formulas: x *satisfies* f if and only if x satisfies the formula "¬(¬g ∨ ¬h)".

Case 5.    f is "g ⊃ h", where g and h are formulas: x *satisfies* f if and only if x satisfies the formula "¬g ∨ h".

Let f be a formula. By *the subset (of Θ) the formula f refers to* (or, alternatively, *the subset (of Θ) the formula f corresponds to*), we mean the set {x: x ∈ Θ and x satisfies f}, denoted as [f].

## References

Black, P.K. (1987). Is Shafer general Bayes? In *Proceedings of the Third Workshop on Uncertainty in Artificial Intelligence*, Seattle, Washington, 2-9.

Collins (1987). COBUILD (COLLINS Birmingham University International Language Database). Collins Publishers, London.

Cox, R.T. (1946). Probability, frequency and reasonable expectation. *American Journal of Physics* 14, 1-13.

Dubois, D. and Prade, H. (1988). Possibilistic and Fuzzy Logics. In *Non-Standard Logics for Automated Reasoning* (P. Smets, E. H. Mamdani, D. Dubois and H. Prade eds.). Academic Press, London.

Dubois, D. and Prade, H. (1990). Updating with belief functions, ordinal conditional functions and possibility measures. In *Proceedings of the Sixth Conference on Uncertainty in Artificial Intelligence*, Cambridge, Massachusetts, 307-315.

Fagin, R. and Halpern, J.Y. (1989). Uncertainty, belief, and probability. In *Proceedings of the Eleventh International Joint Conference on Artificial Intelligence*, Detroit, Michigan, 1161-1167.

Halpern, J.Y. and Fagin, R. (1990). Two views of belief: Belief as generalized probability and belief as evidence. In *Proceedings of the Eighth National Conference on Artificial Intelligence*, American Association for Artificial Intelligence, Boston, Massachusetts, 112-119.

Hsia, Y.-T. (1991a). Characterizing belief with minimum commitment. In *Proceedings of the Twelfth International Joint Conference on Artificial Intelligence*, Sydney, Australia, (to appear).

Hsia, Y.-T. (1991b). Explanations and surprise - a belief-function approach. Technical Report TR/IRIDIA/91-2, IRIDIA, Université Libre de Bruxelles.

Hsia, Y.-T. (1991c). A Belief-Function Semantics for Cautious Nonmonotonicity. Technical Report TR/IRIDIA/91-3, IRIDIA, Université Libre de Bruxelles.

Kyburg, Jr., H.E. (1987). Bayesian and non-Bayesian evidential updating. *Artificial Intelligence* 31, 271-293.

Laskey, K. B. and Lehner, P. E. (1989). Assumptions, beliefs and probabilities. *Artificial Intelligence* 41, 1, 65-77.

Nguyen, H.T. (1978). On random sets and belief functions. *Journal of Mathematical Analysis and Applications* 65, 531-542.

Pearl, J. (1988). *Probabilistic Reasoning in Intelligent Systems: Networks of Plausible Inference*. Morgan Kaufmann Publishers, Inc., San Mateo, California.

Pearl, J. and Geffner, H. (1988). Probabilistic semantics for a subset of default reasoning. TR CSD-870058 (R-94), Cognitive Systems Laboratory, University of California.

Ramsey, F.P. (1931). Truth and probability. In *The Foundations of Mathematics* (Braithwaite, R.B. ed.), Routledge & Kegan Paul, London, 156-198.

Shafer, G. (1976). *A Mathematical Theory of Evidence*. Princeton University Press.

Shafer, G. and Tversky, A. (1985). Languages and designs for probability judgment. *Cognitive Science* 9, 309-339.

Smets, P. (1988). Belief functions. In *Non-Standard Logics for Automated Reasoning* (P. Smets, E. H. Mamdani, D. Dubois and H. Prade eds.). Academic Press, London.

Weaver, W. (1948). Probability, rarity, interest and surprise. *Scientific Monthly* 67, 390-392.

Zadeh, L.A. (1978). Fuzzy sets as a basis for a theory of possibility. Fuzzy Sets and Systems 1, 3-28.